\documentclass[11pt]{article}

\usepackage{amscd,amssymb,stmaryrd}
\usepackage{graphicx}
\usepackage{subcaption}
\usepackage[utf8]{inputenc}
\usepackage[T1]{fontenc}    % use 8-bit T1 fonts
\usepackage[export]{adjustbox}
\usepackage[english]{babel}
\usepackage{cite}
\usepackage{color}
\usepackage{mathtools}
\usepackage{hyperref}
\usepackage{booktabs}
\usepackage{float}

\usepackage{tikz-qtree,amsthm}

\usepackage{diagbox}
\usetikzlibrary{decorations.pathreplacing,calligraphy}
\usepackage{pgfplots}
\pgfplotsset{compat=1.11}
\usepgfplotslibrary{groupplots}

\usepackage{url}
\usepackage{graphicx}

\usepackage{hyperref}       % hyperlinks
\usepackage{url}            % simple URL typesetting
\usepackage{booktabs}       % professional-quality tables
\usepackage{amsfonts}       % blackboard math symbols
\usepackage{nicefrac}       % compact symbols for 1/2, etc.
\usepackage{microtype}      % microtypography
\usepackage{xcolor}         % colors

\usepackage{amsmath}
\usepackage{bm}
\usepackage{multirow}
\usepackage{nicematrix}

\usepackage{authblk}

\title{Time-Series Forecasting, Knowledge Distillation, and Refinement within a Multimodal PDE Foundation Model}

\author[1,*]{Derek Jollie }
\author[2,*]{Jingmin Sun} 
\author[3]{Zecheng Zhang}
\author[4]{Hayden Schaeffer}

\affil[1]{Department of Mathematics, Montana State University, Bozeman, MT 59717.}
\affil[2]{Department of Mathematical Sciences, Carnegie Mellon University, Pittsburgh, PA 15213.}
\affil[3]{Department of Mathematics, Florida State University, Tallahassee, FL 32304.}
\affil[4]{Department of Mathematics, UCLA, Los Angeles, CA 90095.}
\affil[*]{The first two authors contributed equally to this work.}

\date{}

\begin{document}

\maketitle

\begin{abstract}
 Symbolic encoding has been used in multi-operator learning as a way to embed additional information for distinct time-series data. For spatiotemporal systems described by time-dependent partial differential equations, the equation itself provides an additional modality to identify the system. The utilization of symbolic expressions along side time-series samples allows for the development of multimodal predictive neural networks. A key challenge with current approaches is that the symbolic information, i.e. the equations, must be manually preprocessed (simplified, rearranged, etc.) to match and relate to the existing token library, which increases costs and reduces flexibility, especially when dealing with new differential equations.
We propose a new token library based on SymPy to encode differential equations as an additional modality for time-series models. The proposed approach incurs minimal cost, is automated, and maintains high prediction accuracy for forecasting tasks. Additionally, we include a Bayesian filtering module that connects the different modalities to refine the learned equation. This improves the accuracy of the learned symbolic representation and the predicted time-series.
\end{abstract}

\let\thefootnote\relax\footnotetext{The code is available at: \url{https://github.com/JingminSun/prose_v1}.}

\section{Introduction}
Operator learning, initially developed as an application of a universal approximation property in \cite{chen1995universal, chen1993approximations}, aims to approximate maps between functions. Many mathematical and scientific problems can be formulated as the approximation of operators; for instance, forecasting time-series or solving time-dependent partial differential equations (PDEs).
This has made operator learning a crucial tool in computational science and scientific machine learning (SciML) \cite{raissi2019physics, karniadakis2021physics, schaeffer2017learning, schaeffer2017sparse, schaeffer2018extracting, schaeffer2013sparse, zhang2019convergence, lu2021learning, li2020fourier, li2020neural, schaeffer2013low}.
Many deep neural operators (DNOs) \cite{lu2021learning, zhang2023belnet, lin2023b, lu2022comprehensive, li2020fourier, li2022fourier, wen2022u, zhang2024d2no} have been developed and show effectiveness in solving different types of problems relating to time-dependent prediction.
For example in \cite{lu2021learning}, the authors introduced the Deep Operator Network (DeepONet) for approximating the solution map for ordinary differential equations (ODEs) and PDEs.
In \cite{lin2023learning, efendiev2022efficient}, the authors utilize DNOs to predict time-series recursively followed by numerical stabilization.

Though successful in many applications, a key challenge for DNOs is their limited ability to generalize, as they can only handle one operator at a time. 
To address generalization and extrapolation, i.e., the ability to predict new operators and time-series beyond the training interval, multi-operator learning (MOL) \cite{yang2023context, liu2023prose, sun2024towards, sun2024lemon, zhang2024modno, yang2023prompting, yang2024pde, liu2024PROSEFD} has been proposed.
MOL uses a single network structure that is capable of processing data from multiple operators simultaneously. A crucial element of MOL is the way in which the identification of the system is encoded, since this guides the network toward understanding which operator is of interest for a given task, and provides a foundation for extrapolation to new operators.
As a result, MOL networks trained with a diverse dataset and multiple modalities have become an approach for developing PDE foundation models.

Among the proposed MOL approaches, the first two notable contributions include PROSE \cite{liu2023prose, sun2024towards, sun2024lemon, liu2024PROSEFD} 
and ICON \cite{yang2023context, yang2022scalable, yang2024pde}. 
PROSE is the first multimodal PDE foundation model that learns multiple operators and simultaneously predicts the equations that govern the physical system.
It employs a symbolic encoding approach to provide additional information on the PDE of interest by embedding the equations into the feature space.
Additionally, PROSE learns the governing physical system \cite{schaeffer2017learning, schaeffer2017sparse, schaeffer2018extracting, schaeffer2013sparse, zhang2019convergence, sun2020neupde} from the given data simultaneously as it constructs the evaluation operator used in forecasting.
The learned systems are represented as time-dependent PDEs written by symbols and can be used to predict time-series beyond the training time interval.

Symbolic encoding has proven effective in various scenarios, including challenging extrapolation settings \cite{sun2024towards, sun2024lemon}. 
Additionally, PROSE's symbolic encoding can be fine-tuned for downstream tasks and enables zero-shot prediction for new operators \cite{sun2024lemon}.
However, a challenge with symbolic encoding is that PDEs may be written in inconsistent orders and formats. For instance, the expressions $x-1+1+y$ and $y+x$ are mathematically equivalent but would be represented differently in token sequences. 
To address this issue, we propose a standardization approach that utilizes SymPy \cite{sympy} for creating consistent token sequences for the symbolic modality. Our approach demonstrates effectiveness at automatically standardizing the token sequences and improving the encoding process.

\begin{figure}[t]
    \centering
    \includegraphics[width=\linewidth]{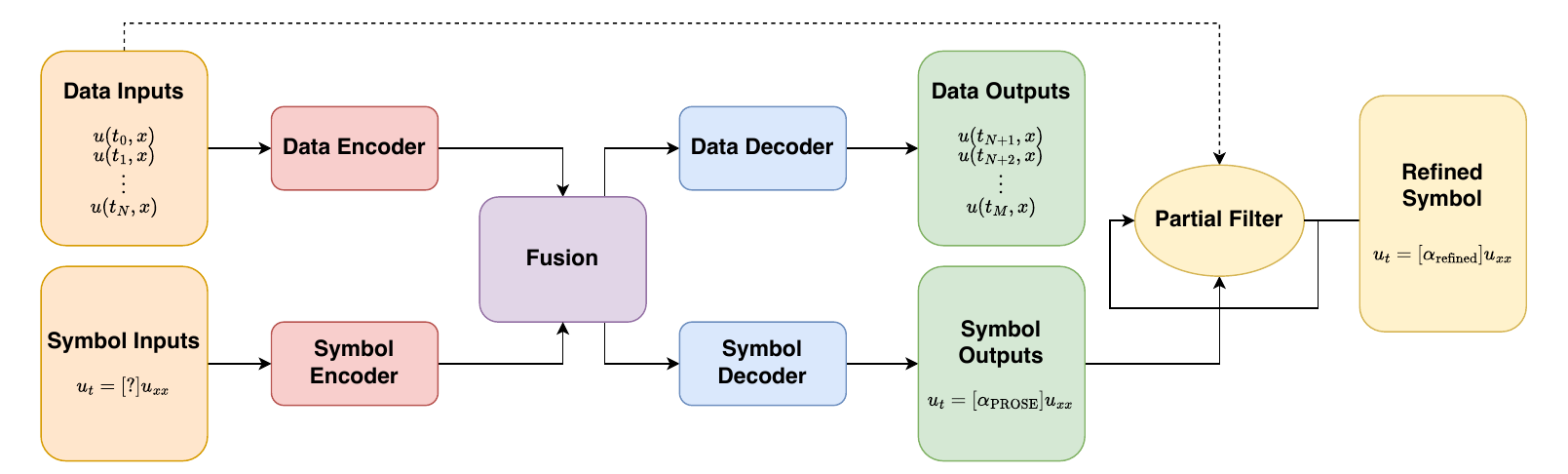}
    \caption{PROSE PDE Foundation Model with Particle Filtering.}\vspace{-1em}
    \label{fig:prose_filter}
\end{figure}

To enhance the prediction accuracy of the physical system, we propose a sequential Monte Carlo (SMC) particle filter module \cite{andrieu2010particle,douc2005comparison,doucet2000sequential,doucet2009tutorial,lin2022theoretical} to refine the learned PDEs within the foundation model framework. Notably, only the coefficients of the PDEs require refinement, as PROSE has demonstrated reliability in correctly identifying the terms in the governing equation, i.e., terms such as $u_{xx}$ and $u_x$. This also holds in the presence of noise or when terms are either missing or mistakenly included. The pipeline of our model is illustrated in Figure~\ref{fig:prose_filter}.

Our main contributions are as follows. 
\begin{itemize}
    \item We propose a new symbolic encoding method that can include a general equation modality. The new method allows the equations to be inputted without a specific format, thus leading to a more flexible model. Compared to the manual standardization methods, the proposed symbol encoding method significantly improves efficiency. 
    \item We examine the ability of PROSE-PDE to generate consistent outputs when given incomplete symbolic inputs.  In the experiments, we test the inclusion of placeholder coefficients on the equations and the addition of incorrect terms in the equations.
    \item A particle filter is introduced to the outputs of the decoders to further refine the learned coefficients, which leads to improved accuracy of the discovered equations. The refined model can be used for stable long-term predictions.
\end{itemize}
The code is available at \url{https://github.com/JingminSun/prose_v1}.

\section{Methods} \label{sec:methods}
Suppose we are given data from $N_{op}$ operators $G_i: U_i \rightarrow V_i$, where $U_i$ and $V_i$ are function spaces. MOL uses a single neural network $\mathcal{G}_{\theta}$ to approximate $G_i$, i.e., $\mathcal{G}_{\theta}(G_i, u )\approx G_i(u)$, where $u\in U_i$ is a given input function for $G_i$, $i = 1, ..., N_{op}$, and $\theta$ is the network parameters. A key component of MOL is the encoding structure used to identify the system of interest as it informs the network of the particular PDE. Our focus is on PDE solution operators, which are crucial for many scientific computing problems. Therefore, we encode the governing equations directly to inform the network of $G_i$. PROSE introduces a symbolic encoding approach for this purpose.  To encode the equations, PROSE represents each equation as a tree with nodes corresponding to operations and leaves to variables or coefficients. This tree is then converted into a sequence (in Polish notation), with each entry consisting of learnable tokens. For example, $\cos(1.5x_1) + x_2^2 -2.6 $ is converted to sequence \texttt{[$+$ cos $\times$ $1.5$ $x_1$ $-$ pow $x_2$ $2$ $2.6$]}, where each entry is a trainable token. This is referred to as the PROSE tree. 

\begin{figure}[b]
\centering
\begin{tikzpicture}[scale=1]
\tikzset{level distance=7mm}
\tikzset{every tree node/.style={align=center,anchor=center, font=\footnotesize}}

\Tree[.\texttt{add} {$u_t$}
    [.\texttt{sub}  
        [.\texttt{mul} {$k$} 
                    [.\texttt{mul} {$u$} {$u_{x}$} ]]
                        [.\texttt{mul} {$\frac{\epsilon}{\pi}$} {$u_{xx}$} ]]]]         
\end{tikzpicture}~~~~~~~~
\begin{tikzpicture}[scale=1]
\tikzset{level distance=7mm}
\tikzset{every tree node/.style={align=center,anchor=center, font=\footnotesize}}

\Tree[.\texttt{add} 
    [.\texttt{mul} {-1} 
        [.\texttt{mul} {$u_{xx}$} {$\frac{\epsilon}{\pi}$} ]]
                    [.\texttt{add} {$u_t$} 
                        [.\texttt{mul} {$u_x$} 
                            [.\texttt{mul} {$k$} {$u$} ]]]]
\end{tikzpicture}
\caption{\textbf{PROSE Tree Examples:} The left tree is an example of a manually standardized PROSE tree for the viscous Burgers' equation $u_t + kuu_x = \frac{\epsilon}{\pi}u_{xx}$.  In the experiments, to generate the randomized trees (or a tree encountered in testing), we randomly switch the order of any branch of the tree with probability $0.5$, leading to different orders of the same symbolic expressions.  The right tree is an example of an altered tree for the same equation.}
\label{fig:tree}
\end{figure}
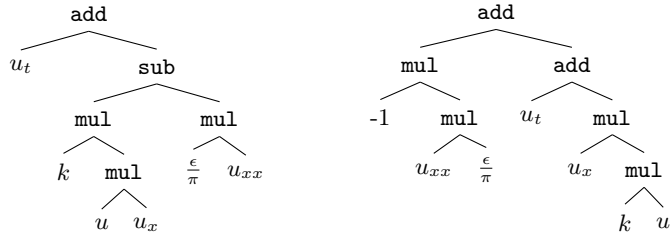

PROSE's symbolic encoding proves effective even in challenging noisy extrapolation settings. However, the equations must be: (1) manually ordered into a particular format and (2) simplified to a standard expression. For example, the equation $u_t - (u_{x})_x  = 0$ would need to be manually formatted as $u_t - u_{xx} = 0$ to ensure that all tokens (operations and variables) fit within the existing library.
This could lead to challenges in the testing phase when an equation is presented with a different order. 
It may become an issue for generalization, as determining the appropriate order when faced with new equations or terms can be difficult. Figure \ref{fig:tree} is an example of the standard order used in PROSE and a possible alternative ordering.
Notably, manually standardizing the tree for new equations with different orders can resolve the problem. However, the manual standardization process is time-consuming and costly.

\begin{figure}[H]
\centering
\begin{tikzpicture}[scale=1]
\tikzset{level distance=10mm}
\tikzset{every tree node/.style={align=center,anchor=center, font=\footnotesize}}

\Tree[.$+$ 
    [.$*$ {1} {$u(x,t)$} 
     [.$\partial$ {(} {$u(x,t)$} {,} {$x$} {)} ]]
        [.$*$ {1} 
            [.$\partial$ {(} {$u(x,t)$} {,} {$t$} {)} ]]
        [.$*$ {$\delta^2$} 
            [.$\partial$ {(} {$u(x,t)$} {,} {(} {$x$} {,} {$3$} {)} {)} ]]]
\end{tikzpicture}
\caption{\textbf{SymPy Tree Example:} KdV equation $uu_x + u_t + \delta^2u_{xxx} = 0$. Here, $\partial(u(x,t),(x,3))$ is used in the tree structure to embed the term $u_{xxx}$ and similarly. Other derivatives are written using this notation.}
\label{fig_sy_tree}
\end{figure}
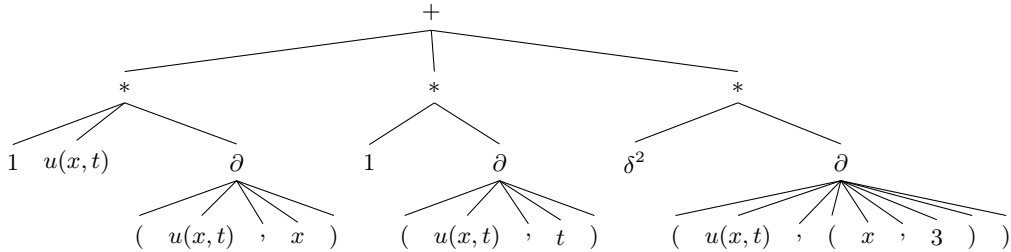

To address these challenges, we first leverage SymPy to unify the expressions. This approach is both fast and cost-effective. 
Note that SymPy processes a mathematical equation using the same symbolic encoding procedure as PROSE but with an added tree-based transformation to simplify equations, i.e., the process is: equation-to-tree, simplified tree, sequence, and SymPy tokens. We refer to this as the \textit{SymPy tree}. Notably, the family of SymPy tokens is larger than the useful tokens needed for encoding PDEs. To address this, we process the SymPy trees by simplifying unnecessary tokens, allowing the updated tokens to be used directly without manual adjustment. For instance, $ u(x,t) $ is tokenized as $``u"$, $``("$, $``x"$, $``,"$, $``t"$, and $``)"$ in SymPy. We simplify this to $``u(x,t)"$, significantly improving both efficiency and accuracy. Figure \ref{fig_sy_tree} is an illustration of the Korteweg–De Vries equation SymPy tree.

\begin{figure}[H]
     \centering
     \includegraphics[width=\linewidth]{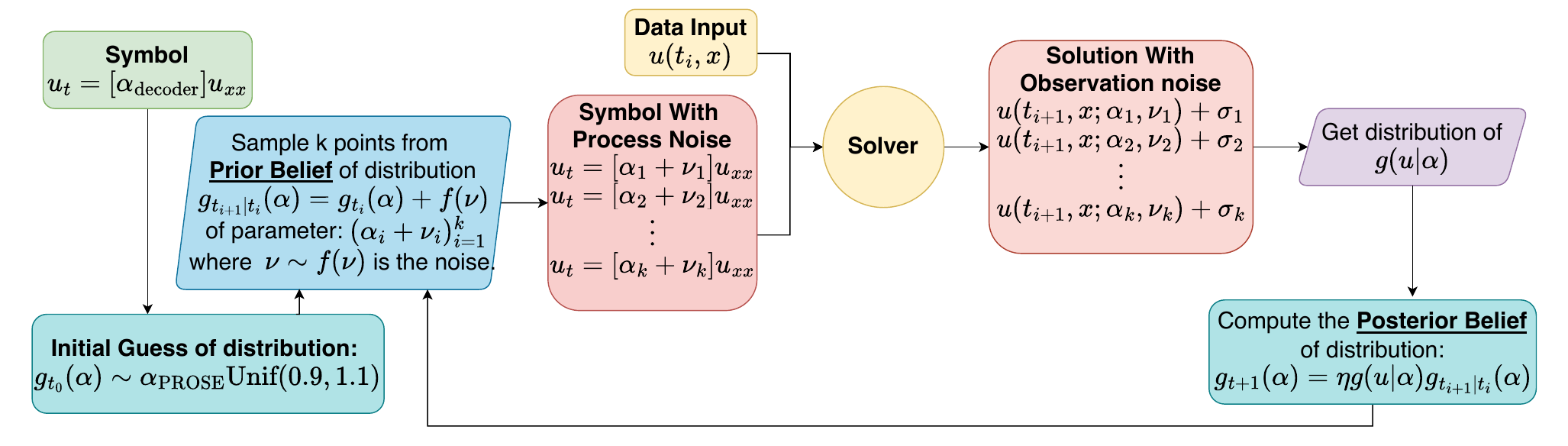}
     \caption{Particle Filter Module: A discretized version of the Bayesian filter process.}
     \label{fig:filter}
 \end{figure}

\textbf{Bayesian Particle Filter}: We propose a refinement module that utilizes the SMC particle filter applied to the symbolic outputs to improve the accuracy of the learned equation. For a coefficient estimated by the symbolic decoder, we first set $\alpha_0 := \alpha_\text{PROSE}$ and the initial distribution $g_0(\alpha_0)$ for the particle filter to be a uniform distribution centered at $\alpha_0$, i.e., $ g_0(\alpha_0) = \text{Unif.}(0.9\alpha_0, 1.1\alpha_0) $. The parameter refinement update rule for $\alpha_k$ is defined as:
$\alpha_k = \alpha_{k-1} + \nu$
where $ \nu \sim \mathcal{N}(0, \epsilon^2) $ is zero-mean Gaussian noise. Using the Chapman-Kolmogorov equation, we compute the prior belief distribution:
\begin{align}
    g_{k|k-1}(\alpha_{k}) = \int_{-\infty}^\infty f_\nu(\nu)g_{k-1}(\alpha_{k-1})d\alpha_{k-1}
    \label{eqn: 6}
\end{align} Bayes' theorem then gives the posterior belief with normal coefficient $\eta$:
\begin{align}
 g_{k}(\alpha_k) = p(\alpha_k | u( t_k,\cdot)) = \eta p(u( t_k,\cdot)| \alpha_k) g_{k|k-1}(\alpha_k).
  \label{eqn: 7}
\end{align}
This process is known as the Bayesian Filtering \cite{chen2003bayesian}, and in practice, we implement it using a SMC particle filter simulation \cite{gustafsson2010particle,doucet2009tutorial,doucet2000sequential,andrieu2010particle, lin2022theoretical}.
The details appear in Section \ref{sec:ParticleFilter} and Figure \ref{fig:filter}.

\subsection{Particle Filter}\label{sec:ParticleFilter}
In this section, we discuss a particle filter algorithm to approximate the distribution for $\alpha$.
We can construct $p(u(\cdot, t_k)| \alpha_{k})$  from the evolution of $u$:
\begin{align}
    u(\cdot, t_k) = H(\alpha_{k}, u(\cdot, t_{k-1})) + \sigma
\end{align}
where $H$ is a (deterministic) numerical scheme for solving the PDE and $\sigma$ is the observation noise. In our case $\sigma \sim f_\sigma(\sigma)$ is sampled from a zero-mean Gaussian distribution with variance $\epsilon^2$.  
Since $\sigma$ is the source of randomness in the dynamics, inserting the distributions into Equation (\ref{eqn: 7}) yields:
\begin{align}
    g_k(\alpha_{k}) = \eta f_\sigma(\sigma)g_{k|k-1}(\alpha_{k}).
    \label{eqn: 9}
\end{align}
To compute this, we utilize a SMC particle method where we generate $M$ particles and approximate \eqref{eqn: 6} by:
\begin{align}
    g_{k|k-1}(\alpha_{k}) \approx \dfrac{1}{M}\sum_{i=1}^M \alpha_{k-1}^{(i)}+\nu =\dfrac{1}{M}\sum_{i=1}^M \alpha_{k|k-1}^{(i)}.
\end{align}
Thus, we take the samples from the initial distribution of $\alpha$ plus the process noise.
 Next, we calculate importance weights \cite{geweke1989bayesian}, which are given by:

\begin{align}
    w_i \propto \dfrac{g_k(\alpha^{(i)})}{ g_{k|k-1}(\alpha_{k}) } = f_{\sigma}(\sigma), \quad \quad    p_i = \dfrac{w_i}{w_0}
    \label{eqn: 11}
\end{align}

with normalization factor $w_0 = \sum_{i=1}^M w_i$.  This allows for discrepancy between the distribution of interest and the distribution from which the samples are drawn.  Finally, we compute the cumulative distribution function for $\alpha$:
\begin{align}
    G_k(\alpha) &= \int_{-\infty}^{\alpha_k}g_k(\zeta)d\zeta \nonumber\\
        &= \int_{-\infty}^{\alpha_k} \eta f_\sigma(\sigma)g_{k|k-1}(\zeta) d\zeta     \nonumber\\
        &\approx \dfrac{1}{\sum_{i=1}^M f_\sigma(\sigma)}\sum_{i=1}^M f_\sigma(\sigma)\mathbb{I}_{(-\infty,\alpha_k)}(\alpha_{k|k-1}^{(i)})\\
        &= \sum_{i=1}^M p_i\, \mathbb{I}_{(-\infty,\alpha_k)}(\alpha_{k|k-1}^{(i)})
\end{align}
where $\mathbb{I}_A$ is the indicator function on the set $A$.  Once this is constructed, we can resample $\alpha_k$ from this new distribution $G_k(\alpha)$, then we repeat this process to construct the next step $\alpha_{k+1}$.

In this work, the particle-based refinement process uses $ M = 500 $ particles and uses $10$ refinement steps. At the last step, we output the mean of $G_{10}(\alpha)$ as the refined parameter, i.e. $\alpha_{\text{refined}} = \frac{1}{M} \sum_{i=1}^M \alpha_{10}^{(i)}.$
The process noise is modeled as a normal distribution with variance equal to $10^{-5}$. Furthermore, the noise introduced by the numerical scheme is also modeled as Gaussian noise, with variance proportional to the initial $ L^2 $-norm of the state $ u $, i.e., $ \epsilon = 0.05 \| u(\cdot,t = 0) \|_2 $.

\section{Numerical Experiments}
We present numerical experiments to demonstrate that our proposed standardized symbol modality enhances prediction performance. 
We investigate five different symbolic encoding settings for PDEs and evaluate the trained model's performance on equations that are not preprocessed, i.e., not simplified or formatted in a specific order.
The five settings used for testing (after pretraining) are:
 (1) PROSE Tree: defined in Section \ref{sec:methods} and \cite{sun2024towards}; (2) Swapping PROSE tree: PROSE tree with randomized ordering for addition and subtraction (with $-1$ multiplied) with probability $0.5$; (3) Noisy Swapping PROSE tree: PROSE tree with random erroneous terms added with probability $0.5$, and the noisy trees are swapped with probability $0.5$; (4) SymPy tree: defined in Section \ref{sec:methods}; and (5) Noisy SymPy tree: SymPy tree with random erroneous terms added with probability $0.5$. 
 We present an example of swapping terms and randomized ordering in Figure \ref{fig:tree}.

\begin{table}[H]
 \setlength{\tabcolsep}{2pt} 
\renewcommand{\arraystretch}{1.2}
    \centering
      \caption{\textbf{PROSE-PDE with Two Modalities.} Noisy: Erroneous terms in the input. Swapping: Rearranged order for terms. $L^2$ and $R^2$ errors are for the data predictions while Symbolic Error and Valid Fraction are metrics for the learned equations, see Appendix~\ref{sec:metrics} for details. PROSE Tree* uses manual formatting and thus not a direct comparison.}
    \scriptsize
    
   \begin{tabular}{cccccc}
         % \hline
   \textbf{Noise}   &  \textbf{Testing Tree Structure}  & \textbf{Relative $L^2$ error} & \textbf{$R^2$ score} &\textbf{Symbolic Error}& \textbf{Valid Fraction}\\
         \hline
        \multirow{3}{*}{ Noise-Free} &PROSE Tree* & 2.18\% & 0.995 & 1.24\%&99.90\%\\\cline{2-6}
         &Swapping PROSE Tree & 3.26\% & 0.983 & 1.43\%&85.94\%\\\cline{2-6}
      &  SymPy Tree &{1.42}\%  &{0.996 }& 1.40\%& {99.95}\%\\
         \hline
        \multirow{2}{*}{Noisy Tree}&   Noisy Swapping PROSE Tree & 4.53\% & 0.968 &2.06\% &76.01\%\\\cline{2-6}
        &Noisy SymPy Tree & {3.81}\%  &  {0.973} & 3.21\% & {83.23}\%\\
         \hline
    \end{tabular}

    \vspace{-1em}
    \label{v1tab_Noisy_2to2}
\end{table}

\begin{figure}[t]
    \centering
\includegraphics[width=\linewidth]{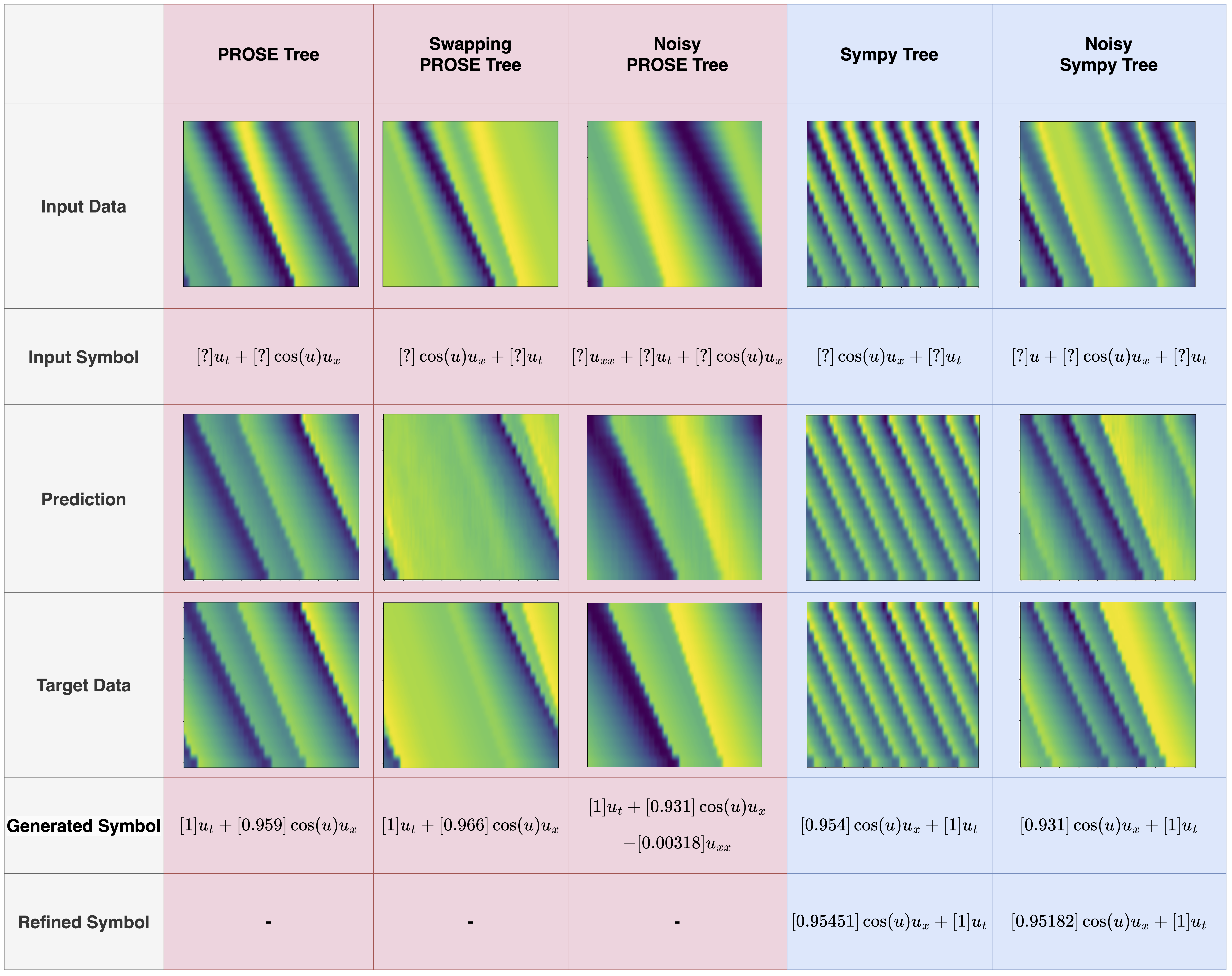}
    \caption{Various examples of the symbolic modality for inviscid conservation law with sine flux. Target equation: $u_t + 0.955 \cos(u)u_x=0$. For PROSE tree, the model is trained for the order $[?]u_t + [?]\cos(u)u_x$, and for SymPy tree, the input expression is automatically uniformed into $ [?]\cos(u)u_x + [?]u_t $. The generated symbols use 3 significant digits while the refinement is a standard float. 
    Notably the SymPy tree removes the erroneous term in prediction.
    See Table \ref{tab:my_label} for error details.}
\vspace{-1em}
    \label{fig:Comparison}
\end{figure}
From Table \ref{v1tab_Noisy_2to2}, we observe that if the order of the terms in the testing equations does not match the training order, the errors increase to $3.26\%$ and $1.43\%$ respectively for the prediction and learned equations. 
In contrast, the SymPy tree achieves the best prediction errors at $1.42\%$ and $1.40\%$, primarily due to the standardization of the format and the token library. This automated process is also faster compared to the manual standardization of the PROSE tree, which resulted in an error of $2.18\%$.
Although the Symbolic Error increases between the Noisy PROSE tree and the Noisy SymPy tree, the valid fraction increases as well, showing more robust knowledge distillation.
To further enhance the prediction accuracy of the physical system and utilize the learned equations to evaluate the time-series, we test the particle filter. Using the model obtained from the previous experiment with the SymPy tree, we randomly select 100 equations from each type for refinement. 
The results are presented in Table \ref{tab:my_label}, with some corresponding predictions illustrated in Figure \ref{fig:Comparison}.

\begin{table}[H]
  \setlength{\tabcolsep}{3pt} 
\renewcommand{\arraystretch}{1.5}
    \centering
        \caption{\textbf{Comparison of Symbolic Modality Errors with and without Particle Filter}. 
    We evaluate PDEs using the learned systems and calculate the Time-Series Errors (see Appendix \ref{sec:metrics}).
    (I)CL: (Inviscid) Conservation Law. }
    \scriptsize
    \begin{tabular}{cccccc}
       \multirow{2}{*}{\textbf{Type of equation}} & \multirow{2}{*}{\textbf{Expression}} &\multicolumn{2}{c}{\textbf{Symbolic Error}}&\multicolumn{2}{c}{\textbf{Time-Series Error}}\\\cline{3-6}
       &&\textbf{Without Filtering} &  \textbf{With filtering}&\textbf{Without Filtering} &  \textbf{With filtering}\\\hline
              Burgers' & $u_t + q_1 ({u^2})_x = q_2u_{xx}$ &1.02\%&0.88\%&1.50\%&1.38\% \\\hline
        Inviscid Burgers' & $u_t + q ({u^2})_x = 0 $&1.11\% & 0.65\% &3.47\%& 2.12\%\\\hline
        CL w. cubic flux & $u_t + q_1 ({u^3})_x = q_2u_{xx}$& 3.07\%& 2.79\%&3.94\%& 3.62\%\\\hline
        ICL w. cubic flux & $u_t + q ({u^3})_x = 0 $& 2.31\%& 1.73\%&3.61\%&2.97\% \\\hline
        ICL w. sine flux & $u_t + q (\sin(u))_x = 0 $&0.50\% & 0.29\%&3.94\%&2.22\% \\\hline
    \end{tabular}
    \vspace{-2em}
    \label{tab:my_label}
\end{table}

\section{Conclusion}
In this work, we propose an automatic equation encoding modality for enhancing the time-series prediction of PDEs within the PROSE foundation model. This approach eliminates the need for costly manual ordering and simplification of PDEs, leading to significant improvements in prediction accuracy. To further refine the governing system learned by PROSE, we include a filter-based module that refines the learned expression. This refinement is possible due to the additional modality in the PDE foundation model. In future work, we will explore alternative refinement techniques to produce accurate and stable long-term predictions.

\subsubsection*{Acknowledgments}
This work was supported in part by NSF 2427558, NSF 2331033, and DE-SC0025440. The authors thank Yuxuan Liu from UCLA for his helpful comments and suggestions.

\clearpage
\newpage
\bibliography{references}
\bibliographystyle{plain}

\newpage
\appendix
\section{Experiment Setup}\label{sec:setup}
\subsection{Dataset}
The dataset utilizes the conservation laws from \cite{sun2024towards}.  To summarize, it consists of 6 families of conservation laws: Inviscid/ viscous Burgers', inviscid/ viscous conservation law with cubic flux, and inviscid/ viscous conservation law with sine flux. The parameters are randomly sampled from $\pm 10 \%$ of the original value and 50 initial conditions leading to 153.6K separate equations used in training. Then 30.72K equations with different parameters are used for testing.  

The initial data sequence is obtained from the PDE dataset using 16 timestamps from $[0,t_f/2]$ ($t_f$ specified per equation) with 128 points for the spatial grid on $[0,x_f]$ for a fixed $x_f$.  Note that a change of variables is used to re-scale and normalize the PDEs so that their solutions reside on a specified interval. We perform data normalization during the training process. Given the data input sequence $\{u(t_i,\cdot)\}_{0\leq i< T_0}$, we compute the mean and standard deviation, which are used to normalized both the input and ground truth label. The loss function is the standard mean squared error in this normalized space.

\subsection{Evaluation Metrics}\label{sec:metrics}

Since we use two modalities, we utilize four evaluation metrics from \cite{sun2024towards}. For metrics on the data, we use the relative $L^2$ error:
$
    \frac{\|u - \tilde{u}\|_2}{\|u\|_2} $,
and the $R^2$ score:
$$R^2 := 1-\frac{\sum_i\|  u_i - \tilde{u}_i\|_2^2}{\sum_i\|u_i - \text{mean} (u_i)\|_2^2}$$
where $u$ is the target, $\tilde{u}$ is the model's prediction, and $i$ is the index for sample.

A \textit{valid} generated expression is considered as the one with true mathematical meanings (i.e. can be decoded into an equation) and with (relative) error less than $100\%$. The percentage of valid expressions are reported and the symbolic error is computed by inputting randomized-coefficient polynomials of the form 
$P(x,t) = (c_0 + c_1t + c_2t^2)(c_3 + c_4x + c_5 x^2 + c_6x^3 + c_7x^4)$ into the learned PDE and the true PDE then taking the relative $L^2$ error between them.  The degree of the polynomials were chosen to avoid the true PDEs from being identically zero. The Time-Series error is the relative $L^2$ error using the prediction generated using the (particle filtered) refined PDE and initial conditions in the input data.

\subsection{Training}
 The models are trained using the AdamW optimizer with batch size of 512 for 30 epochs, where each epoch is 2K steps. The learning rate scheduler is set to have 10\% warmup and a cosine scheduler. We use a learning rate of $10^{-4}$ and weight decay of $10^{-4}$. On a single NVIDIA GeForce RTX 4090 GPUs with 24 GB memory, the training takes about 3.0 hours with PROSE tree and 11.5 hours using SymPy tree. 

\end{document}